\documentclass[a4paper,twoside]{article}

\usepackage{epsfig}
\usepackage{subcaption}
\usepackage{calc}
\usepackage{amssymb}
\usepackage{amstext}
\usepackage{amsmath}
\usepackage{amsthm}
\usepackage{multicol}
\usepackage{pslatex}
\usepackage{apalike}

\usepackage{graphicx}
\usepackage{SCITEPRESS} 

\begin{document}

	\title{Deep Learning Approach to Diabetic Retinopathy Detection}
	\iftrue
	\author{\authorname{Borys Tymchenko\sup{1}\orcidAuthor{0000-0002-2678-7556}, Philip Marchenko\sup{2}\orcidAuthor{0000-0001-9995-9454} and Dmitry Spodarets\sup{3}\orcidAuthor{0000-0001-6499-4575}}
		\affiliation{\sup{1}Institute of Computer Systems, Odessa National Polytechnic University, Shevchenko av. 1, Odessa, Ukraine}
		\affiliation{\sup{2}Department  of Optimal Control and Economical Cybernetics, Faculty of Mathematics, Physics and Information Tecnology, Odessa I.I. Mechnikov National University, Dvoryanskaya str. 2, Odessa, Ukraine}
		\affiliation{\sup{3}VITech Lab, Rishelevska St, 33, Odessa, Ukraine}
		\email{tymchenko.b.i@opu.ua, p.marchenko@stud.onu.edu.ua, dmitry.spodarets@vitechlab.com}
	}
	\fi
	
	\keywords{Deep learning, diabetic retinopathy, deep convolutional neural network, multi-target learning, ordinal regression, classification, SHAP, Kaggle, APTOS.}
	
	\abstract{
		Diabetic retinopathy is one of the most threatening complications of diabetes that leads to permanent blindness if left untreated. One of the essential challenges is early detection, which is very important for treatment success. Unfortunately, the exact identification of the diabetic retinopathy stage is notoriously tricky and requires expert human interpretation of fundus images. Simplification of the detection step is crucial and can help millions of people. Convolutional neural networks (CNN) have been successfully applied in many adjacent subjects, 
		and for diagnosis of diabetic retinopathy itself. However, the high cost of big labeled datasets, as well as inconsistency between different doctors, impede the performance of these methods. In this paper, we propose an automatic deep-learning-based method for stage detection of diabetic retinopathy by single photography of the human fundus. Additionally, we propose the multistage approach to transfer learning, which makes use of similar datasets with different labeling. The presented method can be used as a screening method for early detection of diabetic retinopathy with sensitivity and specificity of 0.99 and is ranked 54 of 2943 competing methods (quadratic weighted kappa score of 0.925466) on APTOS 2019 Blindness Detection Dataset (13000 images). }
	
	\onecolumn \maketitle \normalsize \setcounter{footnote}{0} \vfill
	
	\section{\uppercase{Introduction}}
	\label{sec:introduction}
	\noindent Diabetic retinopathy (DR) is one of the most threatening complications of diabetes in which damage occurs to the retina and causes blindness. It damages the blood vessels within the retinal tissue, causing them to leak fluid and distort vision. Along with diseases leading to blindness, 
	such as cataracts and glaucoma, DR is one of the most frequent ailments, according to the US, UK,
	and Singapore statistics \cite{americanstat,ukstat,singstat}.
	
	DR progresses with four stages:
	\begin{itemize}
		\item \textit{Mild non-proliferative retinopathy}, the earliest stage, 
		where only microaneurysms can occur;
		\item \textit{Moderate non-proliferative retinopathy}, a stage which can be described by losing 
		the blood vessels' ability of blood transportation due to their distortion and swelling with the 
		progress of the disease;
		\item \textit{Severe non-proliferative retinopathy} results in deprived blood supply to the retina 
		due to the increased blockage of more blood vessels, hence signaling the retina for the growing 
		of fresh blood vessels;
		\item \textit{Proliferative diabetic retinopathy} is the advanced stage, where the growth features secreted
		by the retina activate proliferation of the new blood vessels, growing along inside covering of retina in 
		some vitreous gel, filling the eye.
	\end{itemize}
	
	Each stage has its characteristics and particular properties, so doctors possibly could not take some of them into account, and thus make an incorrect diagnosis. So this leads to the idea of creation of an automatic solution for DR detection.
	
\begin{figure}[!h]
		\centering
		\includegraphics[width=.9\linewidth]{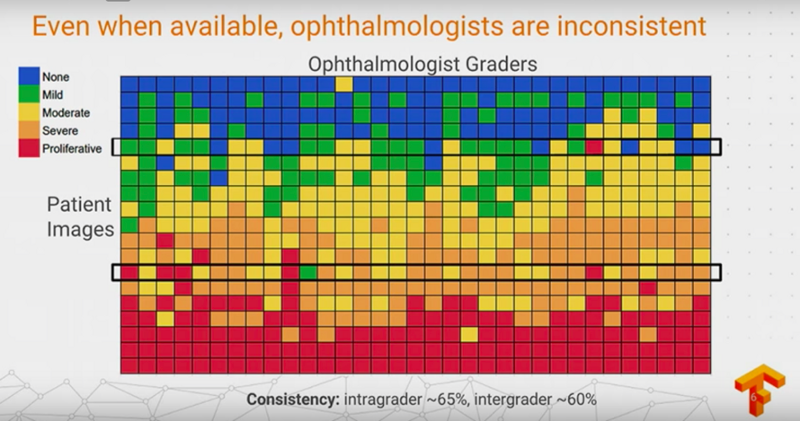}
		\caption{Google showed that ophtalmologists' diagnoses differ for same fundus image. Best viewed in color.}
		\label{fig:incons}
	\end{figure}

	At least 56\% of new cases of this disease could be reduced with proper and timely treatment and monitoring of the eyes \cite{pm}. However, the initial stage of this ailment has no warning signs, and it becomes a real challenge to detect it on the early start. Moreover, well-trained clinicians sometimes 
	could not manually examine and evaluate the stage from diagnostic images of a patient's fundus 
	(according to Google's research \cite{googleopt}, see Figure \ref{fig:incons}). At the same time,  
	doctors will most often agree when lesions are apparent. Furthermore, existing ways of diagnosing are quite inefficient due to their duration time, and the number of ophthalmologists included in patient's problem solution. Such sources of disagreement cause wrong diagnoses and unstable ground-truth for automatic solutions, which were provided to help in the research stage. 
	
	Thus, algorithms for DR detection began to appear. The first algorithms were based on different classical algorithms from computer vision and setting thresholds \cite{abramov,hann,silberman}. 
	Nevertheless, in the past few years, deep learning approaches have proved their superiority over other algorithms in tasks of classification and object detection \cite{pratt}. 
	In particular, convolutional neural networks (CNN) have been successfully applied in many adjacent subjects and for diagnosis of diabetic retinopathy itself \cite{shaoua,pratt}. 
	
	In 2019, APTOS (Asia Pacific Tele-Ophthalmology Society) and competition ML platform Kaggle 
	challenged ML and DL researchers to develop a five-class DR automatic diagnosing solution 
	(APTOS 2019 Blindness Detection Dataset). 
	In this paper, we propose the transfer learning approach and an automatic method for detection of the stage of diabetic retinopathy by single photography of the human fundus. 
	This approach is able to learn useful features even from a noisy and small dataset and could be used as a DR stages screening method in automatic solutions. 
	Also, this method was ranked 54 of 2943 different methods on APTOS 2019 Blindness Detection 
	Competition and achieved the quadratic weighted kappa score of 0.92546.

	\section{\uppercase{Related work}}
	\label{sec:related_work}
	\noindent Many research efforts have been devoted to the problem of early diabetic retinopathy detection. First of all, researchers were trying to use classical methods of computer vision and machine learning to provide a suitable solution to this problem. For instance, Priya et al. \cite{priya} 
	proposed a computer-vision-based approach for the detection of diabetic retinopathy stages using color fundus images. They tried to extract features from the raw image, using the image processing techniques, and fed them to the SVM for binary classification and achieved a sensitivity of 98\%, specificity 96\%, and accuracy of 97.6\% on a testing set of 250 images. Also, other researchers tried to fit other models for multiclass classification, e.g., applying PCA to images and fitting decision trees, naive Bayes, or k-NN  \cite{conde} with best results 73.4\% of accuracy, and 68.4\% for F-measure while using a dataset of 151 images with different resolutions.
	
	With the growing popularity of deep learning-based approaches, several methods that apply CNNs to this problem appeared. Pratt et al. \cite{pratt} developed a network with CNN architecture and data augmentation, which can identify the intricate features involved in the classification task such as micro-aneurysms, exudate, and hemorrhages in the retina and consequently provide a diagnosis automatically and without user input. They achieved a sensitivity of 95\% and an accuracy of 75\% on 
	5,000 validation images. Also, there are other works on CNNs from other researchers \cite{lam,li}.
	It is useful to note that Asiri et al. reviewed a significant amount of methods and datasets available, highlighting their pros and cons \cite{asiri}. Besides, they pointed out the challenges to 
	be addressed in designing and learning about efficient and robust deep-learning algorithms 
	for various problems in DR diagnosis and drew attention to directions for future research.
	
	Other researchers also tried to make transfer learning with CNN 
	architectures. Hagos et al. \cite{hagos} tried to train InceptionNet V3 for 5-class 
	classification with pretrain on ImageNet dataset and achieved accuracy of 
	90.9\%. Sarki et al. \cite{sarki} tried to train ResNet50, Xception Nets, DenseNets and VGG 
	with ImageNet pretrain 
	and achieved best accuracy of 81.3\%. Both teams of researchers used datasets, which were provided by APTOS and Kaggle.
	
	\section{\uppercase{Problem statement}}
	\label{sec:data}
	
	\subsection{Datasets}
	
	The image data used in this research was taken from several datasets. We used an open dataset from Kaggle Diabetic Retinopathy Detection Challenge 2015 \cite{dr2015} for pretraining our CNNs. This dataset is the largest available publicly. It consists of 35126 fundus photographs for left and right eyes of American citizens labeled with stages of diabetic retinopathy:
	
	\begin{itemize}
		\item No diabetic retinopathy (label 0)
		\item Mild diabetic retinopathy (label 1)
		\item Moderate diabetic retinopathy (label 2)
		\item Severe diabetic retinopathy (label 3)
		\item Proliferative diabetic retinopathy (label 4)
	\end{itemize}
	
	In addition, we used other smaller datasets: Indian Diabetic Retinopathy Image Dataset (IDRiD) \cite{idrid}, from which we used 413 photographs of the fundus, and MESSIDOR (Methods to Evaluate Segmentation and Indexing Techniques in the field of Retinal Ophthalmology) \cite{messidor} dataset, from which we used 1200 fundus photographs. As the original MESSIDOR dataset has different grading from other datasets, we used the version that was relabeled to standard grading by a panel of ophthalmologists \cite{messidor_relabeled}.
	
\begin{figure}[!h]
		\centering
		\includegraphics[width=.9\linewidth]{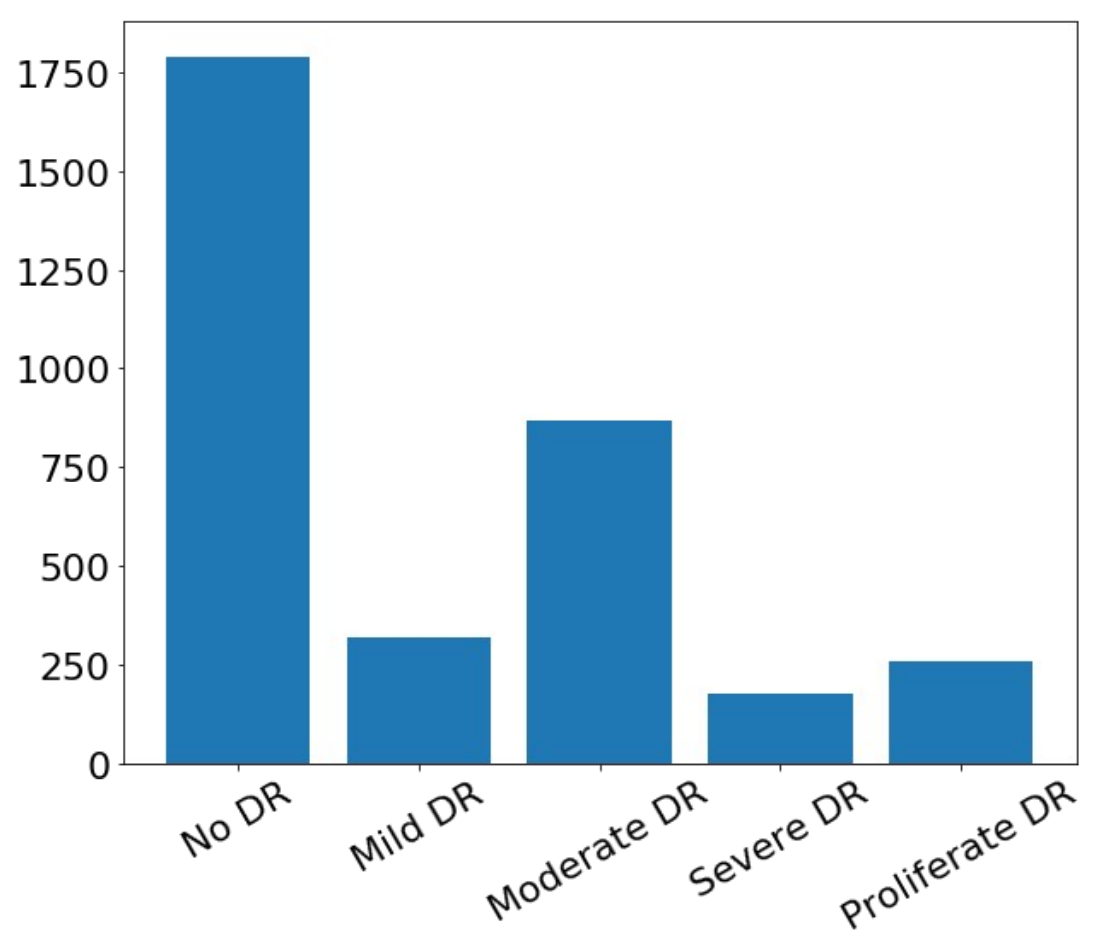}
		\caption{Classes distribution in APTOS2019 dataset.}
		\label{fig:distribution}
	\end{figure}

	As the evaluation was performed on Kaggle APTOS 2019 Blindness Detection (APTOS2019) dataset \cite{aptos2019}, we had access only to the training part of it. The full dataset consists of 18590 fundus photographs, which are divided into 3662 training, 1928 validation, and 13000 testing images by organizers of Kaggle competition. All datasets have similar distributions of classes; distribution for APTOS2019 is shown in Figure \ref{fig:distribution}.
	
	As different datasets have a similar distribution, we considered it as a fundamental property of this type of data. We did no modifications to the dataset distribution (undersampling, oversampling, etc.).
	
\begin{figure}[!h]
		\centering
		\includegraphics[width=.9\linewidth]{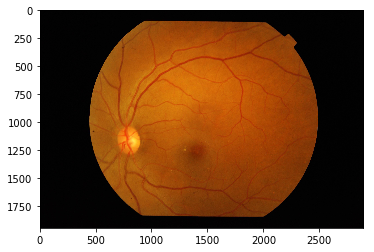}
		\caption{Sample of fundus photo from the dataset.}
		\label{fig:image_samples}
	\end{figure}

	The smallest native size among all of the datasets is 640x480.     
	Sample image from APTOS2019 is shown in Figure \ref{fig:image_samples}.
	
	\subsection{Evaluation metric}
	In this research, we used quadratic weighted Cohen's kappa score as our main metric. Kappa score measures the agreement between two ratings. The quadratic weighted kappa is calculated between the scores assigned by the human rater and the predicted scores. This metric varies from -1 (complete disagreement between raters) to 1 (complete agreement between raters). The definition of $\kappa$ is:
	
	\begin{equation}\label{kappa}
	\kappa = 1- \frac{\sum_{i=1}^{k} \sum_{j=1}^{k}w_{ij}o_{ij}} {\sum_{i=1}^{k} \sum_{j=1}^{k}w_{ij}e_{ij}},
	\end{equation}
	
	where $k$ is the number of categories, $o_{ij}$, and $e_{ij}$ are elements in the observed, and expected matrices respectively. $w_{ij}$ is calculated as following:
	
	\begin{equation}\label{kappa_weights}
	w_{ij} = \frac{(i - j)^2}{(k - 1)^2},
	\end{equation}
	
	Due to Cohen’s Kappa properties, researchers must carefully interpret this ratio. For instance, if we consider two pairs of raters with the same percentage of an agreement, but different proportions of ratings, we should know, that it will drastically affect the Kappa ratio. 
	
	Another problem is the number of codes: as the number of codes grows, Kappa becomes higher. Also, Kappa may be low even though there are high levels of agreement, and even though individual ratings are accurate. All things mentioned above make Kappa a volatile ratio to analyze.
	
	The main reason to use the Kappa ratio is that we do not have access to labels of validation and test datasets. Kappa value for these datasets is obtained by submitting our model and runner's code to the checking system on the Kaggle site. Moreover, we do not have explicit access to images from the test dataset.
	
	Along with the Kappa score, we calculate macro F1- score, accuracy, sensitivity, specificity on holdout dataset of 736 images taken from APTOS2019 training data.
	
	\section{\uppercase{Method}}
	\label{sec:method}
	
	\noindent    The diabetic retinopathy detection problem can be viewed from several angles: as a classification problem, as a regression problem, and as an ordinal regression problem \cite{ordreg}. This is possible because stages of the disease come sequentially. 
	
	\subsection{Preprocessing}
	
	Model training and validation were performed with preprocessed versions of the original images. The preprocessing consisted of image cropping followed by resizing. 
	
\begin{figure}[!h]
		\centering
		\includegraphics[width=.9\linewidth]{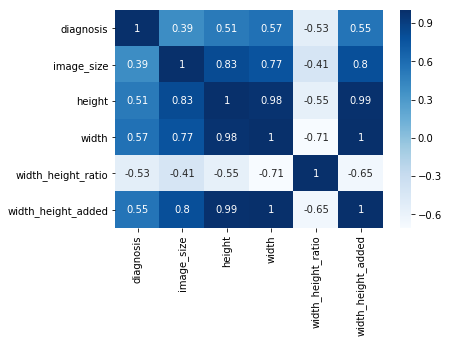}
		\caption{Spurious correlations between meta-features and diagnosis.}
		\label{fig:sp_corr}
	\end{figure}

	Due to the way APTOS2019 was collected, there are spurious correlations between the disease stage and several image meta-features, e.g., resolution, crop type, zoom level, or overall brightness. Correlation matrix is shown in Figure \ref{fig:sp_corr}.
	
	To make CNN be able not to overfit to these features and to reduce correlations between image content and its meta-features, we used a high amount of augmentations. Additionally, as we do not have access to the test dataset both in the competition and in real life, we decided to show as much data variance as possible to models. 
	
	\subsection{Data augmentation}
	
	We used online augmentations, at least one augmentation was applied to the training image before inputting to the CNN. We used following augmentations from Albumentations \cite{albu} library: optical distortion, grid distortion, piecewise affine transform, horizontal flip, vertical flip, random rotation, random shift, random scale, a shift of RGB values, random brightness and contrast, additive Gaussian noise, blur, sharpening, embossing, random gamma, and cutout \cite{cutout}.
	
	\subsection{Network architecture}
	
	We aim to classify each fundus photograph accurately. We build our neural networks using conventional deep CNN architecture, which has a feature extractor and smaller decoder for a specific task (head). 
	
	However, training the encoder from scratch is difficult, especially given the small amount of training data. Thus, we use an Imagenet-pretrained CNNs as initialization for encoder \cite{ternausnet}.
	
	We propose the multi-task learning approach to detect diabetic retinopathy. We use three decoders. Each is trained to solve its task based on features extracted with CNN backbone:
	
	\begin{itemize}
		\item classification head,
		\item regression head,
		\item ordinal regression head.
	\end{itemize}
	
\begin{figure}[!h]
		\centering
		\includegraphics[width=.9\linewidth]{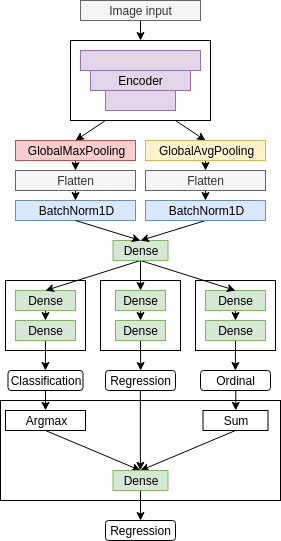}
		\caption{Three-head CNN structure.}
		\label{fig:nn_scheme}
	\end{figure}

	Here, classification head outputs a one-hot encoded vector, where the presence of each stage is represented as 1. 
	Regression head outputs real number in the range $[0, 4.5)$, which is then rounded to an integer that represents the disease stage. 
	For the ordinal regression head, we use the approach described in \cite{ordreg_nn}. 
	Briefly, if the data point falls into category $k$, it automatically falls into all categories from $0$ to $k-1$. So, this head aims to predict all categories up to the target.
	The final prediction is obtained by fitting a linear regression model to outputs of three heads. Neural network structure is shown in Figure \ref{fig:nn_scheme}.
	
	We train all heads and the feature extractor jointly in order to reduce training time. We keep the linear regression model frozen until the post-training stage. 
	
	\subsection{Training process}
	We use a multi-stage training process with different settings and datasets in every stage.
	
	\subsubsection{Pretraining}
	
	We found out that labeling schemes are inconsistent between datasets, so we decided to use the largest one (2015 data) to pretrain our CNNs. Using transfer learning is possible because the natural features of the diabetic retinopathy are consistent between different people and do not depend on the dataset.
	
	In addition, different datasets are collected on different equipment. Incorporation of this knowledge into the model increases its ability to generalize and elevates the importance of natural features by reducing sensitivity to instrument noise.
	
	We initialize feature extractor with weights from Imagenet-pretrained CNN. Heads are initialized with random weights \cite{heinit}. We train a model for 20 epochs on 2015 data with minibatch-SGD and cosine-annealing learning rate schedule \cite{cosinelr}.
	
	Every head is minimizing its loss function: cross-entropy for classification head, binary cross-entropy for ordinal regression head, and mean absolute error for regression head. 
	
	After pretraining, we use encoder weights as initialization for subsequent stages. In our experiments, we observed the consistent improvement of metrics when we substituted weights of heads with random initialization before the main training, so we discard trained heads.
	
	\subsubsection{Main training}
	
	The main training is performed on 2019 data, IDRID, and MESSIDOR combined. Starting with weights obtained in the pretraining stage, we performed 5-fold cross-validation and evaluated models on the holdout set. 
	
	At this stage, we change loss functions for heads: Focal Loss \cite{focalloss} for classification head, binary Focal Loss \cite{focalloss} for ordinal regression head and mean-squared error for regression head. 
	
	We trained each fold for 75 epochs using Rectified Adam optimizer \cite{radam}, with cosine annealing learning rate schedule. To save pretrained weights while new heads are in a random state, we disabled training (froze) of the encoder for five epochs while training heads only. 
	
\begin{figure}[!h]
		\centering
		\includegraphics[width=.9\linewidth]{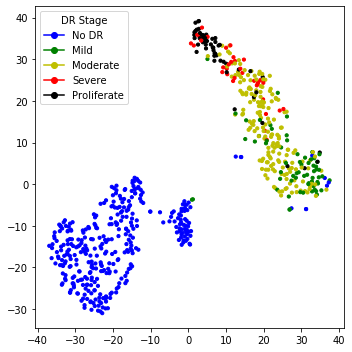}
		\includegraphics[width=.9\linewidth]{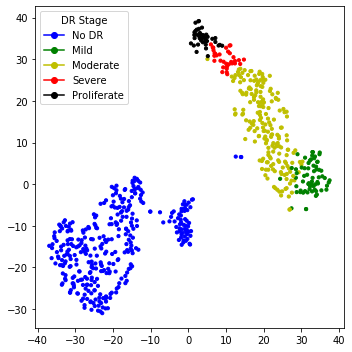}
		\caption{Feature embeddings with T-SNE. Ground truth (top) and predicted (bottom) classes. Best viewed in color.}
		\label{fig:tsne}
	\end{figure}

	During the main training, we monitor separability in feature space generated by the encoder. We generate 2-dimensional embeddings with T-SNE \cite{tsne} and visualize them in the validation phase for manual control of training performance. Figure \ref{fig:tsne} shows T-SNE of embeddings labeled with ground truth data and predicted classes. From the picture, it can be seen that images with no signs of DR are separable with a large margin from other images that have any sign of DR. Additionally, stages of DR come sequentially in embedding space, which corresponds to semantics in real diagnoses.
	
	\subsubsection{Post-training}
	
	In the post-training stage, we only fit the linear regression model to outputs of different heads. 
	
	We found it essential to keep it from updating during previous stages because otherwise, it converges to the suboptimal local minima with weights of two heads close to zero. These coefficients prevent gradients of updating corresponding heads' weights and further discourage network of converging. 
	
	Initial weights for every head were set to $1 / 3$ and then trained for five epochs to minimize mean squared error function.
	
\begin{figure}[!h]
		\centering
		\includegraphics[width=.9\linewidth]{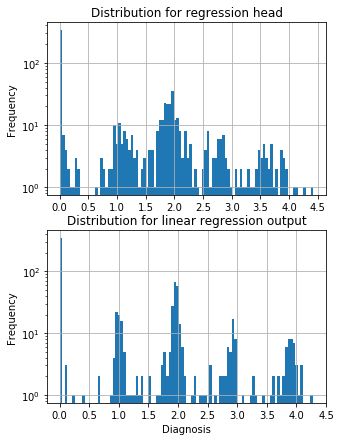}
		\caption{Output distributions for regression head and combination of heads}
		\label{fig:reg_vs_mean}
	\end{figure}

	Difference between prediction distributions of regression head and linear regression outputs is show on Figure \ref{fig:reg_vs_mean}.
	
	\subsubsection{Regularization}
	
	At training time, we regularize our models for better robustness. We use conventional methods, e.g., weight decay \cite{wdecay} and dropout. Also, we penalize the network for overconfident predictions by using label smoothing \cite{lsmoothing}. 
	
	Additionally to label smoothing for classification and ordinal regression heads, we propose label smoothing scheme for linear regression head. It can be used if it is known that underlying targets are discrete. We add random uniform noise to discrete targets:
	
	\begin{align*}
	\label{lsmoothing}
	T_s  &= T + \Delta \\
	\Delta & \sim \mathcal{U}(a, b)
	\end{align*}
	
	Where $T_s$ is smoothed target label, $T$ is the original label, and $\mathcal{U}$ is the uniform distribution. 
	In this case, $-a=b = \frac{T_i - T_{i+1}}{3}$ and $T_i$ $T_{i+1}$ are neighbouring discrete target labels.
	
	Applying this smoothing scheme, we could reduce the importance of wrong labeling.

	\subsubsection{Ensembling}
	
	For final scoring, we ensembled models with 3 encoder architectures at different resolution that scored best on the holdout dataset : EfficientNet-B4 (380x380), EfficientNet-B5 (456x456) \cite{efficientnet}, SE-ResNeXt50 (380x380 and 512x512) \cite{seresnext}.
	
	Our best performing solution is an ensemble of 20 models (4 architectures x 5 folds) with test-time augmentations (horizontal flip, vertical flip, transpose, rotate, zoom). Overall, this scheme generated 200 predictions per one fundus image. These predictions were averaged with a 0.25-trimmed mean to eliminate outliers from possibly overfitted models. A trimmed mean is used to filter out outliers to reduce variance.
	
	We used Catalyst framework \cite{catalyst} based on PyTorch \cite{pytorch} with GPU support. Evaluation of the whole ensemble was performed on Nvidia P100 GPU in 9 hours, processing 2.5 seconds per image.
	
	\section{\uppercase{Results}}
	\label{sec:results}
	
	As experimental results, we provide two tables with metrics, which were mentioned in the Evaluation paragraph. The first table is about results that we have got from local validation without TTA (Table \ref{tab1}), and the second is with TTA (Table \ref{tab2}). 
	
	Our test stage was split into two parts: local testing and Kaggle testing. As we found locally, the ensembling method is the best one, and we evaluated it on Kaggle validation and test datasets. 
	
	On a local dataset of 736 images, ensembling with TTA performed slightly worse than without it. Ensemble with TTA performed better on the testing dataset of 13000 images as it has a better ability to generalize on unseen images. 
	
	Ensembles scored 0.818462/0.924746 validation/test QWK score for a trimmed mean ensemble without TTA and 0.826567/0.925466 QWK score for trimmed mean ensemble with TTA.    
	
	Additionally, we evaluated binary classification (DR/No DR) to check the best model's quality as a screening method (see Tables \ref{tab1} and \ref{tab2}, last row)
	
	The ensemble with TTA showed its stability in the final scoring, keeping consistent rank (58 and 54 of 2943) on validation and testing datasets, respectively.
	
	\begin{table*}[t]
		\centering
		\begin{tabular}{|l||c|c|c|c|c|c|c|c|c|}
			
			\hline
			Model & QWK & Macro F1 & Accuracy & Sensitivity & Specificity\\
			\hline
			EfficientNet-B4 & 0.965 & 0.811 & 0.903 & 0.812 & 0.976\\
			EfficientNet-B5 & 0.963 & 0.815 & 0.907 & 0.807 & 0.977\\
			SE-ResNeXt50 (512x512) & 0.969 & 0.854 & 0.924 & 0.871 & 0.982\\
			SE-ResNeXt50 (380x380) & 0.960 & 0.788 & 0.892 & 0.785 & 0.974\\
			
			Ensemble (mean) & 0.968 & 0.840 & 0.921 & 0.8448 & 0.981\\
			Ensemble (trimmed mean) & 0.971 & 0.862 & 0.929 & 0.860 & 0.983\\ \hline
			Ensemble (trimmed mean, binary classification) & 0.981 & 0.989 & 0.986 & 0.991 & 0.991\\
			\hline
			
		\end{tabular}
		\caption{Results of experiments and metrics tracked, \textbf{without using TTA}.}
		\label{tab1}
	\end{table*}
	
	\begin{table*}[t]
		\centering
		\begin{tabular}{|l||c|c|c|c|c|c|c|c|c|}
			
			\hline
			Model & QWK & Macro F1 & Accuracy & Sensitivity & Specificity\\
			\hline
			EfficientNet-B4 & 0.966 & 0.806 & 0.902 & 0.809 & 0.977\\
			EfficientNet-B5 & 0.963 & 0.812 & 0.902 & 0.807 & 0.976\\
			SE-ResNeXt50 (512x512) & 0.971 & 0.853 & 0.928 & 0.868 & 0.983 \\
			SE-ResNeXt50 (380x380) & 0.962 & 0.799 & 0.899 & 0.798 & 0.976 \\
			Ensemble (mean) & 0.968 & 0.827 & 0.917 & 0.828 & 0.980\\
			Ensemble (trimmed mean) & 0.969 & 0.840 & 0.919 & 0.840 & 0.981\\
			\hline
			Ensemble (trimmed mean, binary classification) & 0.986 & 0.993 & 0.993 & 0.993 & 0.993\\
			\hline
			
		\end{tabular}
		
		\caption{Results of experiments and metrics tracked, \textbf{with using TTA}.}
		\label{tab2}
	\end{table*}

	\section{\uppercase{Interpretation}}
	\noindent In medical applications, it is important to be able to interpret models' predictions. As a good performance of the validation dataset can be a measure to select the best-trained model for production, it is insufficient for real-life use of this model. 
	
	By using SHAP (Shapley Additive exPlanations) \cite{Shapley}, it is possible to visualize features that contribute to the assessment of the disease stage. SHAP unites several previous methods and represents the only possible consistent and locally accurate additive feature attribution method based.
	
	Using SHAP allows ensuring that the model learns useful features during training, as well as uses correct features at inference time. Furthermore, in uncertain cases, visualization of salient features can assist the physician to focus on regions of interest where features are the most noticeable.
	
\begin{figure}[!h]
		\centering
		\includegraphics[angle=90,width=.45\linewidth]{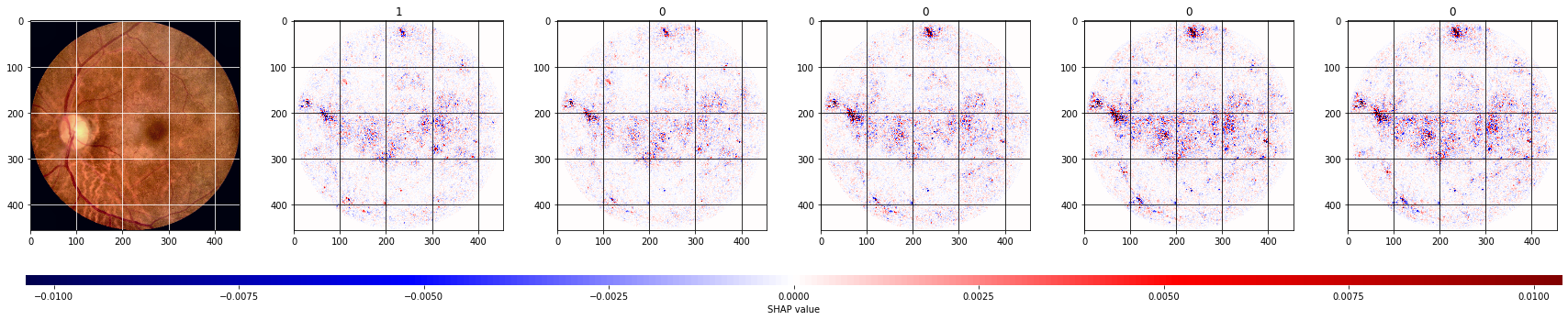}	\includegraphics[angle=90,width=.45\linewidth]{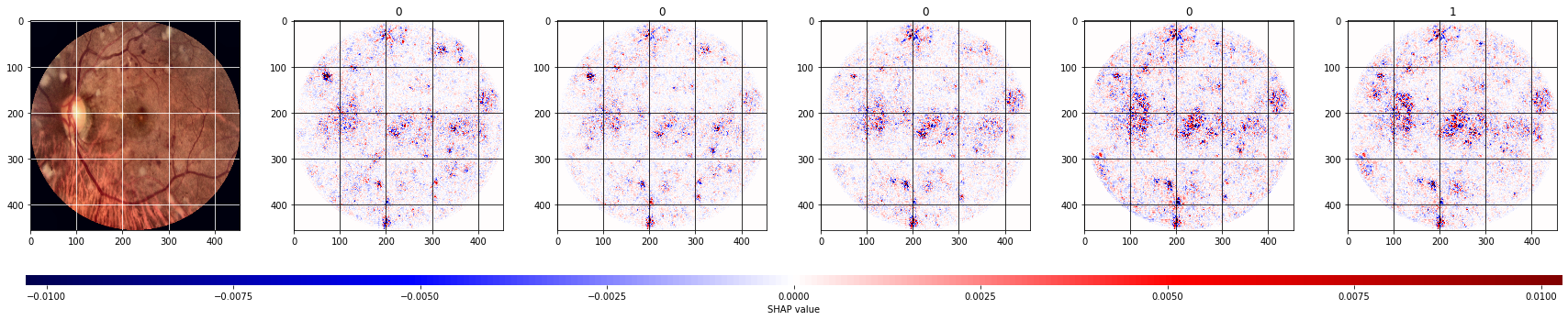}
		\caption{Shap analysis of sample images. Best viewed in color.}
		\label{fig:shap_scheme}
	\end{figure}

	In Figure \ref{fig:shap_scheme}, we show an example visualization of SHAP values for one of the models from the ensemble. Red color denotes features that increase the output value for a given class, and blue color denotes features that decrease the output value for a given class. 
	Overall intensity of the features denotes the saliency of the given region for the classification process. 
	
	\label{sec:interpretation}

	\section{\uppercase{Conclusion}}
	\label{sec:conclusion}
	
	\noindent In this paper, we proposed the multistage transfer learning approach and an automatic method for detection of the stage of diabetic retinopathy by single photography of the human fundus. We have used an ensemble of 3 CNN architectures (EfficientNet-B4, EfficientNet-B5, SE- ResNeXt50) and made transfer learning for our final solution. The experimental results show that the proposed method achieves high and stable results even with unstable metric. The main advantage of this method is that it increases generalization and reduces variance by using an ensemble of the networks, pretrained on a large dataset, and finetuned on the target dataset. The future work can extend this method with the calculation of SHAP for the whole ensemble, not only for a particular network, and with more accurate hyperparameter optimization. 
	Besides, we can do experiments using pretrained encoders on other connected to eye ailments tasks. Also, it is possible to investigate meta-learning \cite{reptile} with these models, but realized that it requires the separate in-depth research.
	
	\bibliographystyle{apalike}
	{\small
		\bibliography{bibliography}}

\end{document}